\documentclass[letterpaper, 10 pt, conference]{ieeeconf}
\IEEEoverridecommandlockouts

\usepackage[table,xcdraw]{xcolor}
\usepackage{threeparttable}
\usepackage{stfloats}
\usepackage{url}
\usepackage{cite}

\usepackage{amsfonts}
\usepackage{algorithm}
\usepackage{array}
\usepackage[caption=false,font=normalsize,labelfont=sf,textfont=sf]{subfig}
\usepackage{pifont}
\usepackage{booktabs}
\usepackage{amsmath}
\usepackage{amssymb}
\usepackage{multirow}
\usepackage{multicol}
\usepackage{subfiles}
\usepackage{pgfplots}
\usepackage{makecell}
\usepackage{bm}
\usepackage{marvosym}
\usepackage{makecell}
\usepackage[colorlinks,linkcolor=red]{hyperref}

\begin{document}


\title{\LARGE \bf LTMSformer: A Local Trend-Aware Attention and Motion State Encoding Transformer for Multi-Agent Trajectory Prediction 
\author{Yixin Yan$^{1}$, Yang Li$^{1}$, Yuanfan Wang$^{1}$, Xiaozhou Zhou$^{1}$, Beihao Xia$^{2}$, Manjiang Hu$^{1}$, Hongmao Qin$^{1}$
}
\thanks{This work is supported by the National Key Research and Development Program of China under Grant 2023YFB2504701 and 2023YFB2504704, and the Open Project Program of Fujian Key Laboratory of Special Intelligent Equipment Measurement and Control under Grant No.FJIES2024KF07. (\textit{Corresponding Author: Yang Li})}
\thanks{$^{1}$ Yixin Yan, Yang Li, Xiaozhou Zhou, Yuanfan Wang, Manjiang Hu, Hongmao Qin are with the College of Mechanical and Vehicle Engineering, Hunan University, Changsha 410082, China. (email: yanyixin@hnu.edu.cn; lyxc56@gmail.com; zxz1340950003@hnu.edu.cn; wangyuanfan@hnu.edu.cn; manjiang\_h@hnu.edu.cn; qinhongmao@vip.sina.com)}
\thanks{$^{2}$ Beihao Xia, School of Electronic Information and Communications, Huazhong University of Science and Technology, Wuhan, Hubei 430074, China, and also with Fujian Key Laboratory of Special Intelligent Equipment Safety Measurement and Control, Fujian Special Equipment Inspection and Research Institute, Fuzhou 350008, China. (email: xbh\_hust@hust.edu.cn)
}
}
\maketitle
\thispagestyle{empty}
\pagestyle{empty}
\begin{abstract}
It has been challenging to model the complex temporal-spatial dependencies between agents for trajectory prediction. As each state of an agent is closely related to the states of adjacent time steps, capturing the local temporal dependency is beneficial for prediction, while most studies often overlook it. Besides, learning the high-order motion state attributes is expected to enhance spatial interaction modeling, but it is rarely seen in previous works. To address this, we propose a lightweight framework, \textit{i.e.}, LTMSformer, to extract temporal-spatial interaction features for multi-modal trajectory prediction. 
Specifically, we introduce a Local Trend-Aware Attention mechanism to capture the local temporal dependency by leveraging a convolutional attention mechanism with hierarchical local time boxes. 
Next, to model the spatial interaction dependency, we build a Motion State Encoder to incorporate high-order motion state attributes, such as acceleration, jerk, heading, etc. To further refine the trajectory prediction, we propose a Lightweight Proposal Refinement Module that leverages Multi-Layer Perceptrons for trajectory embedding and generates the refined trajectories with fewer model parameters. Experiment results on the Argoverse 1 dataset demonstrate that our method outperforms the baseline HiVT-64, reducing the minADE by approximately 4.35\%, the minFDE by 8.74\%, and the MR by 20\%. We also achieve higher accuracy than HiVT-128 with a 68\% reduction in model size.
\end{abstract}

\section{Introduction}
Accurate trajectory prediction is crucial for ensuring safe decision-making in autonomous driving. Temporal-spatial interaction describes how the movements of agents are influenced by both their temporal evolution and interactions with others in space over time \cite{ngiam2021scene,8957246}. Modeling the temporal-spatial interaction dependencies between agents has shown huge benefits in enhancing the accuracy of trajectory predictions \cite{huang2022survey}. However, it is challenging to model temporal-spatial interactions due to the dynamic and uncertain nature of agent behaviors in complex environments \cite{feng2023macformer}. \par

\begin{figure}
    \centering
    \includegraphics[width=1\linewidth]{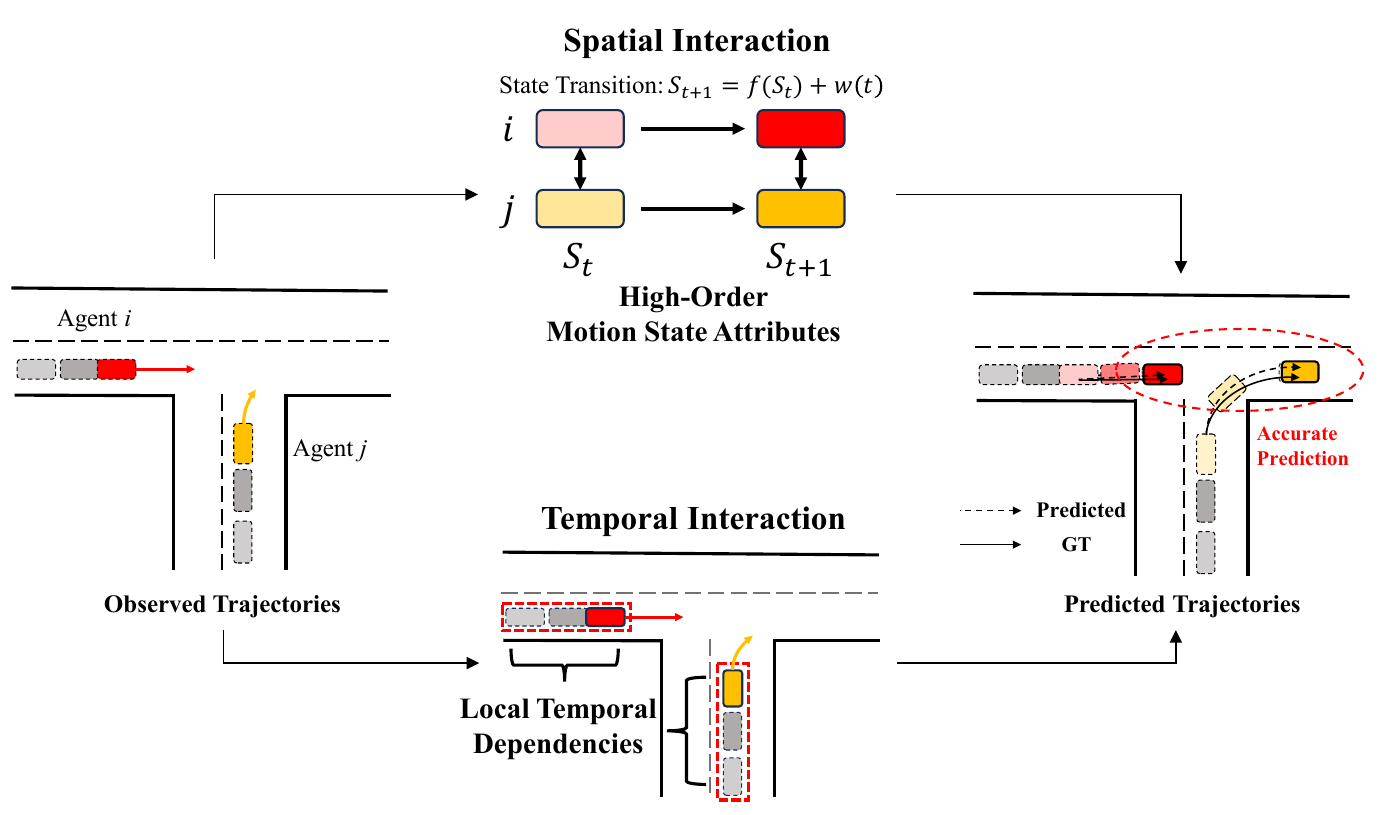}
    \caption{Predicted trajectories are generated from the observed trajectories through interaction modeling. By learning local motion trends in the temporal dimension and encoding high-order motion state attributes in the spatial dimension, the temporal-spatial dependencies are captured. This enhances the accuracy of trajectory predictions while maintaining alignment with the ground truth.}
    \label{fig:1}
\end{figure}
Many studies have been proposed for trajectory prediction, mainly including the physics-based models \cite{barth2008,batz2009recognition,brannstrom2010model} and learning-based models \cite{schreier2016integrated,bahram2015game,he2019probabilistic,li2019dynamic,alahi2016social,chandra2019traphic,ma2019trafficpredict,gao2020vectornet,park2023leveraging,xu2022groupnet,jia2023hdgt,ye2021gsan, azadani2024novel,zhou2022hivt,nayakanti2023wayformer,tang2024hierarchical}.
The physics-based models \cite{barth2008,batz2009recognition,brannstrom2010model} mainly use dynamics models or kinematics models of agents for trajectory prediction. These models are simple, but they often fail to account for complex social interactions that are crucial in multi-agent systems for accurate trajectory prediction. The learning-based approaches generate future trajectories based on the observed trajectory data and contextual information \cite{schreier2016integrated,bahram2015game,he2019probabilistic,li2019dynamic,alahi2016social,chandra2019traphic,ma2019trafficpredict,gao2020vectornet,park2023leveraging,xu2022groupnet,jia2023hdgt,ye2021gsan, azadani2024novel,zhou2022hivt,nayakanti2023wayformer,tang2024hierarchical}. To model the social interactions between agents, Long Short-Term Memory (LSTM)-based frameworks \cite{alahi2016social, chandra2019traphic, ma2019trafficpredict} have been widely proposed, while LSTM is hard to capture the long-range dependencies. 
Graph Neural Networks (GNNs) \cite{gao2020vectornet,park2023leveraging,xu2022groupnet,jia2023hdgt,ye2021gsan} and Transformer-based models \cite{azadani2024novel, zhou2022hivt,nayakanti2023wayformer,tang2024hierarchical} have recently been proposed to learn the temporal-spatial dependencies between multiple agents.  
However, in trajectory prediction tasks, the current state is closely related to the states of adjacent time steps, and Transformer-based models typically process the entire sequence simultaneously, making it hard to learn the local temporal dependencies between adjacent time steps \cite{zhou2022hivt,nayakanti2023wayformer,tang2024hierarchical}.
Besides, capturing the high-order motion patterns of agents can help improve trajectory prediction accuracy, while the GNN-based methods often overlook the high-order motion state attributes of agents, such as acceleration, jerk, etc., in spatial interaction modeling \cite{zhou2022hivt,nayakanti2023wayformer,tang2024hierarchical}.

Accurately capturing local temporal dependencies and high-order motion state attributes of agents is essential for improving temporal-spatial interaction modeling. Therefore, we propose a lightweight trajectory prediction framework, LTMSformer, which enhances the extraction of temporal-spatial interaction features to improve trajectory prediction accuracy, as shown in Fig. \ref{fig:1}. Specifically, we capture temporal interactions through hierarchical local time boxes and trend-aware attention. Motion state encoder is designed to encode high-order motion state attributes to strengthen spatial interaction modeling. Additionally, we introduce a lightweight proposal refinement module to leverage the fusion of temporal-spatial interaction features for trajectory refinement.
The main contributions are summarized as follows:
\begin{itemize}
    \item We propose a lightweight trajectory prediction framework, LTMSformer, which models the temporal-spatial dependencies via local trend-aware attention mechanism and the high-order motion state encoding.
    \item The local trend-aware attention captures the local temporal motion trends of agents by leveraging convolutional attention with hierarchical local time boxes of different sizes. Simultaneously, the motion state encoder enhances spatial interactions by leveraging the high-order motion state attributes. 
    \item We propose a lightweight proposal refinement module to generate refined trajectories based on the fusion of temporal-spatial interaction features, which enhances the accuracy and consistency of predicted trajectories with fewer parameters. Experimental results show quantitative and qualitative superiority.
\end{itemize}

The structure of this paper is organized as follows: Section \ref{section2} reviews related works. Section \ref{section3} presents the overall framework, problem formulation, and key components of LTMSformer. Section \ref{section4} provides the experimental setup and results. Section \ref{section5} concludes the paper.

\section{Related Work}
\label{section2}
\noindent
\textbf{Temporal-Spatial Interaction Modeling.}
Modeling temporal-spatial interactions is beneficial in predicting the future trajectories of multiple agents \cite{zhou2022hivt,nayakanti2023wayformer}. Previous studies often use machine learning methods for interaction modeling \cite{schreier2016integrated,bahram2015game,he2019probabilistic,li2019dynamic}. For instance,
\cite{bahram2015game} utilizes a Deep Belief Network to model the interactive factors between agents and predict their motion states and vehicle maneuvers through game theory. \cite{li2019dynamic} incorporates multiple predictive features, including observed vehicle states, road structures, and vehicle social interactions, to predict the likelihood of each maneuver. LSTM and its variants have also been widely used for interaction modeling. Social LSTM \cite{alahi2016social} introduces a social pooling strategy for modeling agent interactions. However, LSTM-based models \cite{alahi2016social, chandra2019traphic,ma2019trafficpredict} face challenges in capturing long-term interaction dependencies.\cite{gao2020vectornet,park2023leveraging,xu2022groupnet,jia2023hdgt,ye2021gsan} have been used to model agent interactions with graphs. For example, VectorNet \cite{gao2020vectornet} employs a fully connected graph to model
the higher-order interactions.
GroupNet \cite{xu2022groupnet} learns a multi-scale interaction hypergraph using an association matrix to capture the complex interactions among agents. 
Transformers have also been used to model the dynamic temporal-spatial interactions \cite{azadani2024novel,zhou2022hivt,tang2024hierarchical,nayakanti2023wayformer, PuITSC}.
HiVT \cite{zhou2022hivt} extracts the temporal-spatial features of agents through an agent-centric local scene structure that is invariant to translation.
Overall, most studies ignore the local temporal dependency and only consider relative positions for spatial interaction modeling. Different from \cite{zhou2022hivt, nayakanti2023wayformer, tang2024hierarchical},  we propose the Local Trend-Aware Attention mechanism to capture the local temporal dependencies within locally adjacent time steps and use the Motion State Encoder to encode the high-order motion attributes for learning spatial dependencies. This approach can efficiently capture the temporal-spatial interaction features among agents.

\noindent \textbf{Trajectory Refinement Frameworks.} 
Trajectory refinement is the process of improving an initially predicted trajectory to achieve higher accuracy through multi-stage trajectory prediction.
\cite{liu2024laformer, wang2023prophnet} use the first-stage trajectories as anchors to refine trajectory offsets in the second stage. 
MTR\cite{shi2022motion} utilizes target points as queries and applies attention layers to aggregate contextual information for iterative trajectory refinement. 
Multipath++\cite{varadarajan2022multipath++} employs learned anchor embeddings, replacing the static anchors, to align with the modes of trajectory distribution and predict multiple possible trajectories.
However, existing refinement methods often rely on large-scale networks or iterative strategies that would increase model parameters. Different from \cite{shi2022motion,choi2023r,varadarajan2022multipath++}, we propose a Lightweight Proposal Refinement Module that utilizes several MLPs instead of gated recurrent units or attention layers for trajectory embeddings. This approach enables precise single-pass refinement with fewer model parameters while improving prediction accuracy.
\section{Methodology}
\label{section3}
\subsection{Overall Framework}
\begin{figure*}
    \centering
    \setlength{\abovecaptionskip}{-0.2cm}
    \includegraphics[width=0.98\linewidth]{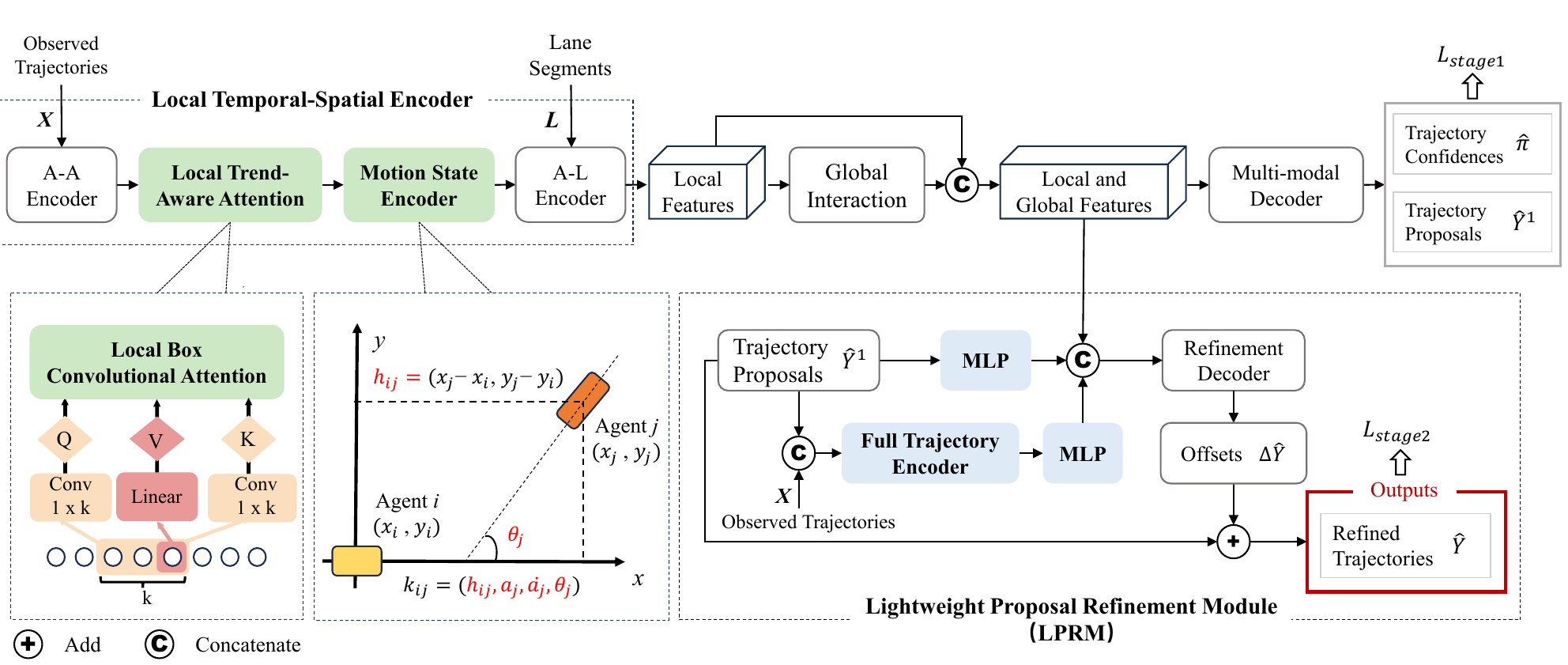}
    \caption{Based on HiVT \cite{zhou2022hivt}, our method introduces three additional components, including Local Trend-Aware Attention Mechanism (LTAA), Motion State Encoder (MSE), and Lightweight Proposal Refinement Module (LPRM). LTAA and MSE are utilized to capture local temporal-spatial interaction dependencies, and LPRM is used to refine the trajectory proposals. Here, A-A and A-L represent Agent-Agent and Agent-Lane interactions, respectively.}
    \label{fig:2}
\end{figure*}
Fig. \ref{fig:2} illustrates the architecture of LTMSformer, which consists of two stages. In the first stage, we present the Local Temporal-Spatial Encoder, which consists of an Agent-Agent Encoder, a Local Trend-Aware Attention (LTAA), a Motion State Encoder (MSE), and an Agent-Lane Encoder. Different from HiVT \cite{zhou2022hivt}, we utilize LTAA and MSE to capture local temporal-spatial interaction dependencies. Specifically, LTAA effectively extracts multi-scale local temporal motion trends by leveraging a convolutional attention mechanism with hierarchical local time boxes, while MSE incorporates high-order motion state
attributes from surrounding agents to
model the spatial interaction dependencies.
Then, Global Interaction aggregates these local features to embed the global social interactions. The Multi-modal Decoder then generates multi-modal trajectory predictions. In the second stage, the Lightweight Proposal Refinement Module (LPRM) refines these multi-modal trajectories by integrating both local and global temporal-spatial interaction features.


\subsection{Problem Formulation}
In this study, we use \( \bm{X} =  \left\{\bm{x}_i^t\right\} \) to represent the observed trajectory of agent \( i \), where $i \in \left\{1, 2, \dots, N \right\}$, and $t \in \left\{1, 2, \dots, T_o\right\}$, $N$ is the number of agents, \( \bm{x}_i^t \) is the coordinates of agent \( i \) at time step \( t \), and \( T_o \) is the observed time horizon. Similarly, we use \( \bm{\hat{Y}} = \left\{\bm{\hat{y}}_i^t\right\}\) to represent the predicted trajectory of agent \( i \), where $t \in \left\{{1, 2, \dots, T_p}\right\}$, \( \bm{\hat{y}}_i^t \) is the predicted coordinates of agent \( i \) at time step \( t \), and \( T_p \) is the predicted time horizon.
Then, We utilize the observed trajectories $\bm{X}$ and lane segment embeddings \( \bm{L} \) to predict future trajectories as follows:
\begin{equation}
{f_p}: (\bm{X}, \bm{L}) \to \bm{\hat{Y}},
\end{equation}
where \( {f _p}\) is the trajectory prediction model. 
Similar to \cite{zhou2022hivt}, we vectorize agent trajectories and lane segments at each local region to ensure translation invariance. Specifically, for the agent \( i \)-th at time step \( t \), the trajectory is represented as $\Delta \bm{p}_i^t = \bm{p}_i^t - \bm{p}_i^{t-1}$, where \( t \in \left\{1, 2, \dots, T_o \right\}\) and \( \bm{p}_i^t \in \mathbb{R}^2 \) represents the position \( (p_{i,x}^t, p_{i,y}^t) \) of the \( i \)-th agent at time step \( t\). We describe the relative states of other agents \( j\) with respect to a central agent \( i \) at the time step \( t \) as $\Delta \bm{p}^t_{ij} = \Delta \bm{p}^t_j - \Delta \bm{p}^t_i$. 
We select the trajectory \( \Delta \bm{p}_i^{T_o}= \bm{p}_i^{T_o}-\bm{p}_i^{T_o-1}\) of the central agent \( i \) at the current time step $T_o$ as the reference vector for the local region, its orientation \( \theta_i\) is then used to define a rotation matrix $\bm{R}_i^{T_o}$ to align all trajectory segments accordingly. The rotated trajectory features are computed as:
\begin{equation}
\bm{h}_i^{t} = \bm{R}_i^{T_o} \Delta \bm{p}_i^t,\quad \bm{h}_{ij}^t = \bm{R}_i^{T_o} \Delta \bm{p}_{ij}^t,
\end{equation}
where \( \bm{R}_i^{T_o} \in \mathbb{R}^{2 \times 2} \) is the rotation matrix parameterized by \( \theta_i\). Similarly, lane segments are also converted into the agent-centric coordinate frame. For a lane segment \( \bm{L} \), its attribute $\bm{P}_L$ is defined as $\bm{P}_L= \bm{P}_1^L - \bm{P}_0^L$, where $\bm{P}_0^L, \bm{P}_1^L \in \mathbb{R}^2$ are the start and end positions of lane segment, respectively. 
\subsection{{Local Temporal-Spatial Encoder
}}
Local Temporal-Spatial Encoder includes Agent-Agent Encoder, LTAA, MSE, and Agent-Lane Encoder are shown in Fig. \ref{fig:2}.
\begin{figure}
    \centering
    \includegraphics[width=1\linewidth]{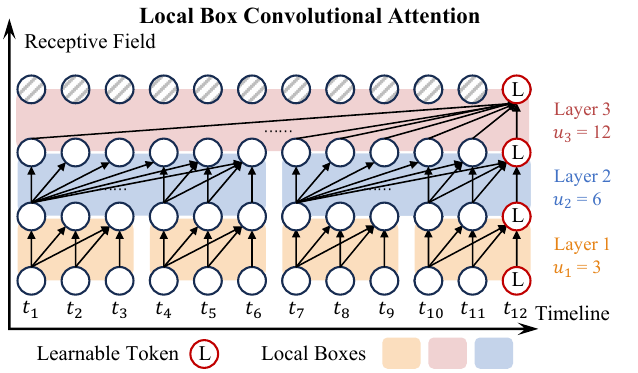}
    \caption{Illustration of Local Box Convolutional Attention with local boxes of sizes 3, 6, and 12 at each layer. In this study, we use local boxes of sizes 3, 7, and 21 at each layer, depending on the length of the sequence in the Argoverse 1 motion prediction dataset.}
    \label{fig:3}
\end{figure}
\subsubsection{\textbf{Agent-Agent Encoder}}
This module aims to capture the interaction features between the central agent and its neighboring agents over the time steps from $t=1$ to $T_{o}$. Given the inputs $\bm{h}_i^t$, $\bm{h}_j^t$, and $\bm{h}_{ij}^t$, the embeddings are computed as follows:
\begin{equation}
\bm{s}_i^t = \phi_{\text{center}}(\bm{h}_i^t) , \quad \bm{s}_{ij}^t = \phi_{\text{nbr}}(\bm{[h}_j^t, \bm{h}_{ij}^t]),
\label{eq:4}
\end{equation}
where $\phi_{\text{center}(\cdot)}$ and $\phi_{\text{nbr}(\cdot)}$ are MLP blocks. Following HiVT\cite{zhou2022hivt}, we adopt the multi-head attention (MHA) mechanism to represent the spatial interaction features \( \bm{c}_i^t \) among agents:
\begin{equation}
\bm{c}_i^t = \text{MHA}(\bm{s}_i^t, \bm{s}_{ij}^t),
\label{eq:9}
\end{equation}
where \( \bm{c}_i^t \in \mathbb{R}^{d_h} \), ${d_h}$ is the model dimension.
Finally, we concatenate the output \( \bm{c}_i^t \) from time step 1 to \( T_{\text{o}} \) as \( \bm{z}_i = \{\bm{c}_i^t\}_{t=1}^{T_{\text{o}}} \), where \( \bm{z}_i \in \mathbb{R}^{T_{\text{o}} \times d_h} \), and put it into the subsequent LTAA module.
\subsubsection{\textbf{Local Trend-Aware Attention Mechanism}}
LTAA aims to capture the local temporal motion trends of agents in observed time steps, thereby enhancing the ability to model agent social interactions in the temporal dimension.
Like \cite{zhou2022hivt}, we append a learnable token $\bm{z}^{T_o+1} \in \mathbb{R}^{d_h}$ at the end of the input sequence and add learnable positional embeddings to all tokens. These tokens are then stacked into a matrix $\bm{S}_i \in \mathbb{R}^{(T_o+1) \times d_h}$, which is fed into the LTAA module.
Next, the input sequence is processed through a causal $1 \times k$ convolution layer with BatchNorm to generate queries $\bm{Q}_i^{LTAA}$ and keys $\bm{K}_i^{LTAA}$, where $k$ is the convolution kernel size in the temporal dimension, a linear transformation is then applied to obtain values $\bm{V}_i^{LTAA}$:
\begin{equation}
\begin{aligned}
\bm{Q}_i^{LTAA} &= \text{BatchNorm(conv}(\bm{z}_i)), \\
\bm{K}_i^{LTAA} &= \text{BatchNorm(conv}(\bm{z}_i)), \\
\bm{V}_i^{LTAA} &= \bm{W}_V^{LTAA} \bm{z}_i,
\end{aligned}
\end{equation}
where $\bm{W}_V^{LTAA}$ is a learnable matrix. 
We apply causal $1 \times k$ convolutions within each box to ensure the temporal trend features at the current time step depend on the current and previous $k-1$ time steps. 
Considering that adjacent time steps typically exhibit stronger correlations than distant ones, we apply local box convolutional attention (denoted as \(\text{LMHA}\), see Fig. \ref{fig:3}) within non-overlapping local boxes to capture temporal motion trends, formulated as:
\begin{equation}
\bm{b}_i^{u_v} = \text{LMHA}(\bm{Q}_i^{LTAA}, \bm{K}_i^{LTAA}, \bm{V}_i^{LTAA}),
\end{equation}
where \(\bm{b}_i^{u_v}\) represents the embedded local temporal motion trend for the $v$-th layer, with box size of \(u_v\). To capture multi-scale local temporal motion trends while preserving the large receptive field of standard $\text{MHA}$, we progressively increase the local box size across different layers of local box convolutional attention from bottom to top. Specifically, LTAA consists of three cascaded layers, each operating at different local box sizes, denoted as \(\bm{b}_i^{u_1}\), \(\bm{b}_i^{u_2}\), and \(\bm{b}_i^{u_3}\). The outputs of these layers are cascaded to form the final unified representation $\bm{B}_i = \text{Cascade}(\bm{b}_i^{u_1}, \bm{b}_i^{u_2}, \bm{b}_i^{u_3})$. Compared to multi-head attention, LTAA reduces parameter complexity by restricting attention computations to localized matrices, thereby improving computational efficiency.
Similar to BERT \cite{lee2018pre}, we use the GELU activation function in the feedforward layer to enhance the model performance.


\subsubsection{\textbf{Motion State Encoder} }
In trajectory prediction, the high-order motion state attributes of agents play a crucial role in determining their future trajectories. The ability to model and understand the dynamic interactions between agents is essential for accurate prediction. To achieve this, we have designed the MSE, which is deployed after the LTAA module to capture the high-order motion state attributes and further enhance the spatial interactions of the agents.
MSE encodes motion state attributes in a vector $\bm{k}_{ij}$, defined as:
\begin{equation}
    \bm{k}_{ij} = [\bm{h}_{ij}, \bm{a}_{j}, \dot{\bm{a}_{j}}, \theta_{j}],
\end{equation}
where $\bm{h}_{ij}$ represents the relative positions of agents, while $\bm{a}_{j}$, $\dot{\bm{a}_{j}}$ and $\theta_{j}$ correspond to the acceleration, jerk, and heading of neighboring agents, respectively. Specifically, the relative positions and heading angles are spatial information, while acceleration and jerk are higher-order motion information.
We employ an MLP denoted as $\phi_\text{m}(\cdot)$ to embed high-order motion state attributes $\bm{k}_{ij}$, and implement MHA with $\bm{B}_{i}$ from LTAA module to generate a temporal-spatial embedding $\bm{e}_i$:
\begin{equation}
    \bm{e}_i = \text{MHA}(\bm{B}_i, \phi_\text{m}(\bm{k}_{ij})),
\end{equation}
where $\bm{B}_i$ serves as the query, while $\phi_\text{m}(\bm{k}_{ij})$ serves as both the key and value in the MHA process. The MSE enables the model to effectively integrate motion state information, which is then fed into the subsequent agent-lane encoder module, improving the accuracy of trajectory prediction.

\subsubsection{\textbf{Agent-Lane Encoder}}
To capture the interaction between agents and surrounding lanes, we introduce the encoding of local lane information:  
\begin{equation}  
\bm{z}_i^L = \phi_{\text{lane}} \left( \left[ \bm{R}_i^{T_o} \bm{P}_L,\bm{R}_i^{T_o} \left(\bm{P}_0^L - \bm{p}_i^{T_o}\right), a^L \right] \right),  
\end{equation}  
where $\phi_{\text{lane}}(\cdot)$ is an MLP encoder for lane segment features, and $a^L$ denotes the semantic attributes of the lane segment. For more details, see HiVT \cite{zhou2022hivt}.
We can obtain the local interaction embedding as follows:
\begin{equation}
    \psi^\text{local}_i = \text{MHA}(\bm{e}_i, \bm{z}_i^L),
\end{equation}
where $\psi^\text{{local}}_i$ is final local interaction embedding, $\bm{e}_i$ is the query, both key and value are $\bm{z}_i^L$.

\subsection{Global Interaction}
In this module, we use an MLP $\phi_{\text{global}}(\cdot)$ to encode the relative positions and angular differences between agents, generating interaction embeddings for each agent pair. Then, the pairwise embedding $\bm{g}_{ij}$ is defined as:  
\begin{equation}
  \bm{g}_{ij} = \phi_{\text{global}} \left(\left[ \bm{h}_{ij}^{T_o},\cos(\Delta \theta_{ij}), \sin(\Delta \theta_{ij})\right] \right),
\end{equation}
where $\Delta \theta_{ij}$ represents the relative angle between agents $i$ and $j$. Then, we leverage the MHA mechanism to combine local interaction embedding $\psi^\text{{local}}_i$ and the pairwise embedding $\bm{g}_{ij}$, formulated as:   
\begin{equation}
    \psi^\text{{global}}_i = \text{MHA}\left(\psi^\text{{local}}_i, [ \psi^\text{{local}}_j,\bm{g}_{ij}] \right),
\end{equation}
where $\psi^\text{{global}}_i$ is the global interaction embedding. After this step, we could integrate rich information about surrounding agents and lane segments.

\subsection{Multi-modal Decoder}
We use three separate MLP blocks $\phi_\text{{d}}(\cdot)$, to process $\psi^\text{{local}}_i$ and $\psi^\text{{global}}_i$. These blocks output the concatenation of the predicted future position $\bm{\hat\mu}_{i,m}$, scale $\bm{\hat\beta}_{i,m}$, and the associated probability $\bm{\hat\pi}_{i,m}$ for each agent $i$ and the trajectory mode $m$:
\begin{equation}
    \{(\bm{\hat\mu}_{i,m}, \bm{\hat\beta}_{i,m}), \bm{\hat{\pi}}_{i,m}\}_{m=1}^{M} = \phi_\text{{d}}([\psi^\text{local}_{i}, \psi^{\text{global}}_i]),
\end{equation}
where $M$ is the number of modes, we define the distribution of the stage-one output trajectories as a mixture of Laplace distributions, formulated as:
\begin{equation}
    {\bm{\hat{Y}}_i^{\text{1}}} = \sum_{m=1}^{M} \bm{\hat\pi}_{i,m} \text{Laplace}(\bm{\hat\mu}_{i,m}, \bm{\hat\beta}_{i,m}),
\end{equation}
where ${\bm{\hat{Y}}_i^{\text{1}}}$ represents the predicted trajectory distribution of agent \(i\) at the first stage.
\subsection{Lightweight Proposal Refinement Module}
The LPRM refines the initial trajectory predictions using lightweight MLP layers, effectively reducing parameter complexity while enhancing prediction accuracy.

\subsubsection{\textbf{Proposal Embedding}}
The initial trajectory prediction serves as the proposal, providing a foundation for subsequent refinement in the second stage. We use an MLP \( \phi_\text{p}(\cdot) \) to process \( \hat{Y}_i^{\text{1}} \), formulated as:
\begin{equation}
       \psi^\text{P}_i = \phi_\text{p}(\bm{\hat{Y}}_i^{\text{1}}),
\end{equation}
where $\psi^\text{P}_i$ is the trajectory proposal embedding.
\subsubsection{\textbf{Consist Embedding} }
The full trajectory \( \bm{Y}_i^{\text{full}} \) represents as \( \bm{Y}_i^{\text{full}}=[\bm{X},\bm{\hat{Y}}_i^{\text{1}}] \),  
is processed using a two-layer MLP with residual connections, denoted as \( f_{\text{full}} \), and formulated as:
\begin{equation}
  Y^f_{i} = f_{\text{full}}(\bm{Y}_i^{\text{full}}),
\end{equation}
where ${Y}_i^f$ is the refined full trajectory embedding.
Then, we use a three-layer MLP \( \phi_{\text{cons}(\cdot)} \) to process \( {Y}_i^f \) for consistency processing, capturing physical constraints and temporal-spatial regularities:
\begin{equation}
    \psi_i^\text{cons} = \phi_{\text{cons}}({Y}_i^f),
\end{equation}
where \( \psi_i^\text{cons} \) is the consistency trajectory embedding that enhances trajectory continuity and physical plausibility.

\subsubsection{\textbf{Proposal Refinement Decoder} }
We utilize a three-layer MLP \( \phi_{\text{refine}}(\cdot) \) to process the joint embedding \( \psi^\text{refine}_i=\text{Concat}(\psi_i^\text{cons},\psi^\text{P}_i,\psi^\text{local}_i,\psi^\text{global}_i) \) and compute the trajectory offset $\Delta \bm{\hat{Y}}_i$, as expressed by:  
\begin{equation}
\Delta \bm{\hat{Y}}_i = \phi_{\text{refine}}(\psi^\text{refine}_i),
\end{equation}
\begin{equation}
\bm{\hat{Y}}_{i} = \bm{\hat{Y}}_i^{\text{1}}+ \Delta \bm{\hat{Y}}_i,
\end{equation}
where \( \bm{\hat{Y}}_{i} \) is the stage-two predicted trajectory of agent \( i \).
LPRM exclusively utilizes MLPs for a lightweight design, significantly reducing computational overhead while enhancing the accuracy, continuity, and physical plausibility of trajectory predictions.
\subsection{Loss Function} 
The loss function includes two-stage losses to optimize trajectory diversity and accuracy:
\begin{equation}
L_{all} = L_{\text{stage1}} + \lambda_1 L_{\text{stage2}},
\end{equation}
where \(L_{all}\) is the total loss function, \( \lambda_1 \) is a hyperparameter controlling the importance of the second stage, $ L_{\text{stage1}}$ and $ L_{\text{stage2}}$ are the loss functions for the first and second stages, respectively.
\( L_{\text{stage1}} \) aims to promote the diversity of multi-modal trajectories and consists of classification loss \( L_{\text{cls}} \) and regression loss \( L_{\text{reg}} \):
\begin{equation}
    L_{\text{stage1}} = L_{\text{cls}} + L_{\text{reg}}.
\end{equation}

Similar to \cite{zhou2022hivt}, \( L_{\text{cls}} \) uses the cross-entropy loss to measure the difference between the predicted probabilities and the target distribution. \( L_{\text{reg}} \) uses negative log-likelihood to encourage the diversity of multi-modal trajectories.
\( L_{\text{stage2}} \) aims to further refine the trajectory and improve the accuracy of the prediction using the Smooth L1 loss function:
\begin{equation}
L_{\text{stage2}} = \frac{1}{NT_{\text{p}}} \sum_{i=1}^{N} \sum_{t=T_{\text{o}} + 1}^{T_\text{o}+T_{\text{p}}} \text{SmoothL1}(\bm{\hat{Y}}_{i}^t - \bm{G}_{i}^t),
\end{equation}
\begin{equation}
\text{SmoothL1}(x) =
\begin{cases}
0.5x^2, & \text{if } |x| < 1, \\
|x| - 0.5, & \text{otherwise.}
\end{cases}
\end{equation}
where $\bm{G}_{i}^t$ represents the ground truth, $\text{SmoothL1}(\cdot)$ is the Smooth L1 loss function, which enhances trajectory prediction accuracy by effectively balancing error penalization.

\begin{table*}[htbp]
    \centering  
    \caption{Performance of Single Model Without Ensembling on the Argoverse 1 Validation and  Test Sets.}
    \label{tab:results}
    \setlength{\tabcolsep}{6pt} 
    \renewcommand{\arraystretch}{1.5}
    \begin{threeparttable}
    \tabcolsep=0.40cm
    \begin{tabular}{c|c|ccc|ccc|c}
        \hline
        \multirow{2}{*}{Model} & \multirow{2}{*}{Year} & \multicolumn{3}{c|}{Validation Set} & \multicolumn{3}{c|}{Test Set} & \multirow{2}{*}{\#Param↓ }\\
        & & minADE↓ & minFDE↓ & MR↓ & minADE↓ & minFDE↓ & MR↓ & \\
        \hline
        DenseTNT\cite{gu2021densetnt} & 2022 & 0.73 & 1.05 & 0.10 & 0.8817 & 1.2815 & \underline{0.1258} & 1103k \\
        LTP\cite{wang2022ltp} & 2022 & 0.78 & 1.07 & - & 0.8335 & 1.2955 & 0.18566 & 1100K \\
        \rowcolor{gray!20}  
        HiVT-64$\dagger$\cite{zhou2022hivt} & 2022 & 0.69 & 1.03 & 0.10 & 0.8306 & 1.3053 & 0.1503 & \textbf{662k} \\
        \rowcolor{gray!20}  
        HiVT-128$\dagger$\cite{zhou2022hivt} & 2022 & \textbf{0.66} & \underline{0.96} & \underline{0.09} & \underline{0.7993} & 1.2320 & 0.1369 & 2529k \\
        FRM\cite{park2023leveraging} & 2023 & \underline{0.68} & 0.99 & - & 0.8165 & 1.2671 & 0.1430 & - \\
        Macformer-S\cite{feng2023macformer} & 2023 & - & - & - & 0.8490 & 1.2704 & 0.1311 & 879k \\
        TMF\cite{azadani2024novel} & 2024 & - & - & - & \textbf{0.7980} & \underline{1.2200} & 0.1360 & - \\
        HVTD\cite{tang2024hierarchical} & 2024 & \underline{0.68} & 1.02 & 0.10 & 0.8200 & 1.2700 & 0.1500 & - \\
        \rowcolor{gray!20}  
        LTMSformer (ours) & 2025 & \textbf{0.66} & \textbf{0.94} & \textbf{0.08} & \underline{0.7993} & \textbf{1.1932} & \textbf{0.1254} & \underline{789k} \\
        \hline
    \end{tabular}
    \begin{tablenotes}
      \footnotesize
      \item[*] Best in bold. The second best is underlined.
      The symbol “$\dagger$” denotes the model obtained locally using the official checkpoint.
    \end{tablenotes}
    \end{threeparttable}
\end{table*}
\section{Experiments}
\label{section4}

\subsection{Experimental Setup}
\textbf{Dataset.} We evaluate LTMSformer on the Argoverse 1 motion prediction dataset.
The dataset contains 323,557 real-world driving scenarios, each lasting 5 seconds. The data is represented in a 2D Bird's Eye View format, with centroid locations sampled at 10 Hz. Moreover, it includes high-definition maps with lane centerlines, traffic directions, and intersection annotations.
The dataset is split into training (205,942 scenes), validation (39,472 scenes), and test (78,143 scenes) sets. The test set provides 2 seconds of observed data, requiring models to predict the next 3 seconds of motion.

\textbf{Metrics.} 
Following \cite{zhou2022hivt}, models on the Argoverse 1 dataset are evaluated using standard motion prediction metrics, including minimum Average Displacement Error (minADE), minimum Final Displacement Error (minFDE), and Miss Rate (MR). These metrics are used to evaluate the predicted six potential trajectories for each agent.

\textbf{Implementation Details.} All experiments are conducted on an RTX 4090 GPU for 64 epochs using the AdamW optimizer. The hyperparameters are as follows: a batch size of 32, an initial learning rate of $5 \times 10^{-4}$, a weight decay factor of $1 \times 10^{-4}$, a dropout rate of 0.1, the number of predictive modes $M$ set to 6, the loss weight term $\lambda_1$ is 5, the number of LTAA layers is set to 3 with local boxes size B = \{3,7,21\}, respectively. The learning rate is dynamically adjusted using a cosine annealing schedule. The radius for all local regions is set at 50 meters. We select the modality with the lowest minFDE as the best prediction result.
We do not utilize any tricks such as data augmentation techniques or ensemble methods in this work. The experiments are conducted based on a model with 64 hidden units.

\subsection{Comparative Study}
We compare the performance of LTMSformer with several state-of-the-art models on the Argoverse 1 validation set and test set, as presented in Table \ref{tab:results}. We focus solely on a single model without ensembling or other tricks, ensuring a fair comparison with other single models. Our model, LTMSformer, achieves the best results across the minADE, minFDE, and MR metrics in the validation set. Specifically, compared to the base HiVT-64, LTMSformer demonstrates a reduction in prediction errors, lowering the minADE by approximately 4.35\% (from 0.69 to 0.66), the minFDE by about 8.74\% (from 1.03 to 0.94), and the MR by 20\% (from 0.10 to 0.08).
Similarly, on the Argoverse 1 test set, LTMSformer maintains superior performance, as shown in Table \ref{tab:results}. It achieves the lowest values for minFDE and MR among all methods. In particular, compared to the baseline HiVT-64, LTMSformer reduces the minADE by 3.77\% (from 0.8306 to 0.7993), the minFDE by 8.59\% (from 1.3053 to 1.1932), and the MR by 16.57\% (from 0.1503 to 0.1254). Our method also achieves higher accuracy than HiVT-128 with a 68\% reduction in model size, 2529k versus 789k.

\begin{figure*}
    \centering
    \setlength{\abovecaptionskip}{-0.2cm}
    \includegraphics[width=1\linewidth]{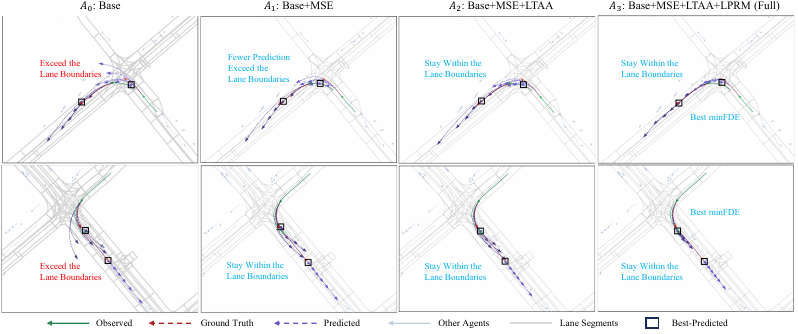}
    \caption{Visualization of the ablation study result. It is carried out by incrementally adding MSE, LTAA, and LPRM, highlighting the contribution of each component to the overall trajectory prediction accuracy. $A_0$ is our base model HiVT-64, $A_3$ is LTMSformer (Base+MSE+LTAA+LPRM).}
    \label{fig:4}
\end{figure*}

 \begin{figure*}
    \centering
\setlength{\abovecaptionskip}{-0.2cm}
\includegraphics[width=0.96\linewidth]{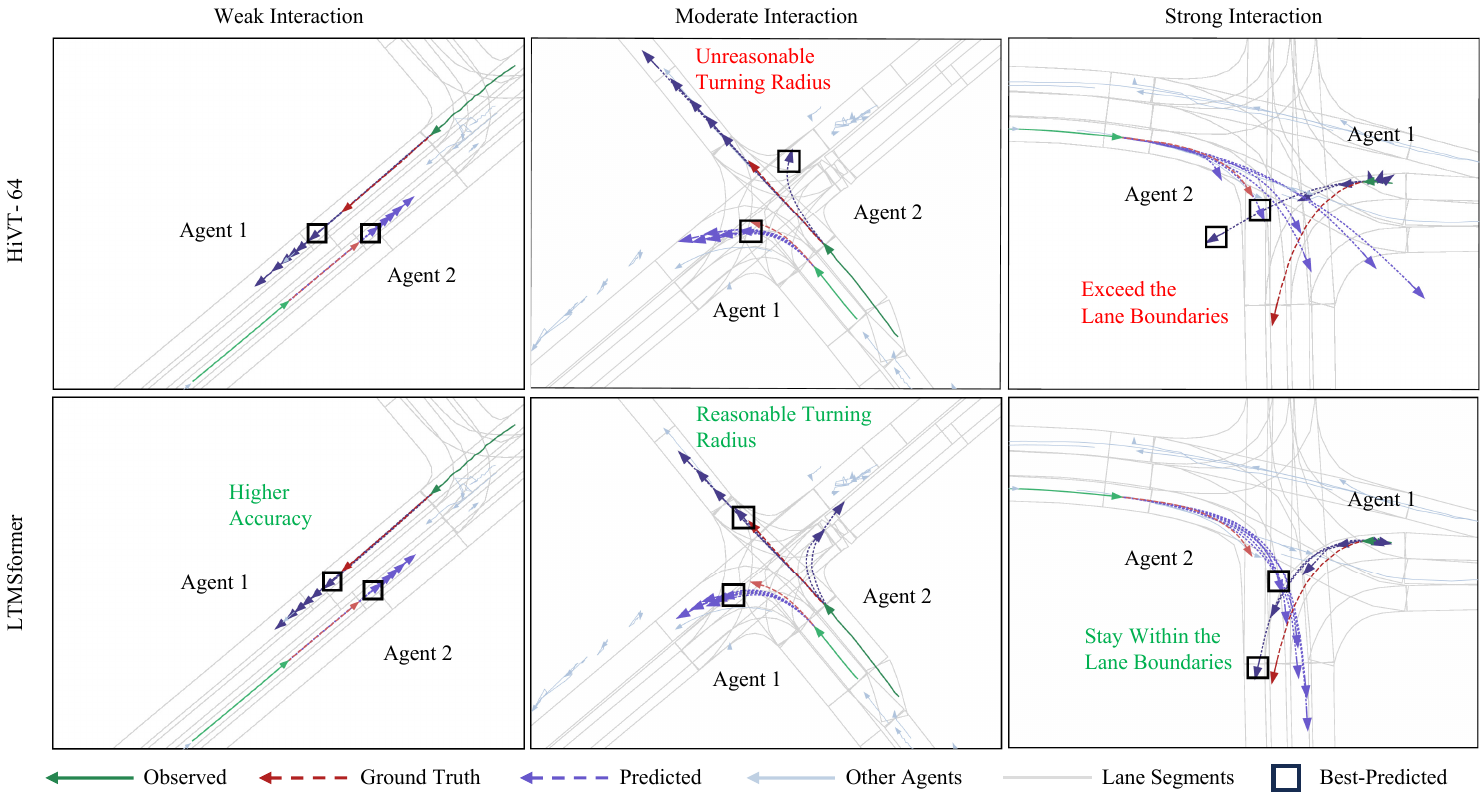}
    \caption{Visualization results of LTMSformer and HiVT-64 under weak, moderate, and strong interaction scenarios. LTMSformer consistently achieves a lower minFDE metric and generates final points closer to the ground-truth endpoints. In the weak interaction scenario, our method achieves higher accuracy; in the moderate interaction scenario, HiVT-64 has an unreasonable turning radius, while our method does not; and in the strong interaction scenario, LTMSformer stays within the lane boundaries, while HiVT-64 exceeds the lane boundaries, showcasing our method provides more reasonable and accurate predictions.}
    \label{fig:5}
\end{figure*}

\begin{table}[htbp]
\caption{Ablation Study on Argoverse 1 Validation Set.}
 \setlength{\tabcolsep}{4.5pt} 
\renewcommand{\arraystretch}{1.5}
\centering
\begin{tabular}{c|ccc|ccc}
\hline
   Model & MSE & LTAA & LPRM & minADE$\downarrow$ & minFDE$\downarrow$ & MR$\downarrow$ \\
\hline
 $A_0$ & - & - & - & 0.687 & 1.030 & 0.103 \\
 $A_1$ & \checkmark & - & - & 0.673 & 1.001 & 0.098 \\
 $A_2$& \checkmark & \checkmark &-  & 0.672 & 1.000 & 0.096  \\
 $A_3$ & \checkmark & \checkmark & \checkmark & \textbf{0.661} & \textbf{0.945} & \textbf{0.082} \\
\hline
\end{tabular}
\label{tab:ablation}
\end{table}

\subsection{Ablation Study}

The ablation study on the Argoverse 1 validation set systematically evaluates performance by incrementally adding each component. 
The inclusion of MSE, LTAA, and LPRM consistently and significantly improves prediction accuracy, as evidenced by the progressive reductions in minADE, minFDE, and MR, as shown in Table \ref{tab:ablation}. 
Model $A_1$ (Base+MSE) shows improvements over the base model $A_0$ (HiVT-64), with further gains observed when adding LTAA in model $A_2$ (Base+MSE+LTAA). Finally, model $A_3$ (Base+MSE+LTAA+LPRM), which includes all components, achieves the best performance, with the lowest minADE (0.661), minFDE (0.945), and MR (0.082), highlighting the effectiveness of the full LTMSformer framework. 

The visualization results from the ablation experiments are illustrated in Fig. \ref{fig:4}. Initially, in $A_0$, the predicted trajectories deviate significantly from the lane and even exceed its boundaries. Introducing the MSE in $A_1$ leads to noticeably improved prediction results, with trajectories closer to the ground truth and a substantial reduction in endpoint errors, achieving fewer predictions that exceed the lane boundaries. This demonstrates the critical role of the MSE module in capturing motion state attributes by encoding the motion state attributes of agents. Next, $A_2$ incorporates LTAA, ensuring that all predicted trajectories stay within lane boundaries while aligning trajectory trends more closely with the ground truth. This demonstrates LTAA's capability to capture motion trends in the temporal dimension. Finally, $A_3$ integrates the LPRM based on $A_2$, which not only ensures that all predicted trajectories stay within lane boundaries but also minimizes the overall error between the predictions and the ground truth. This highlights the essential role of LPRM in refining trajectory proposals and enhancing temporal-spatial consistency. In summary, the gradual incorporation of MSE, LTAA, and LPRM consistently enhances model performance in dynamic traffic interaction scenarios, as demonstrated by both qualitative and quantitative analyses.


\begin{table}
\caption{Sensitivity Study. $\lambda_1$: Parameter in Loss Function.}
\centering
\setlength{\tabcolsep}{17pt} 
\renewcommand{\arraystretch}{1.5}
\begin{tabular}{c|ccc}
\hline
$\lambda_1$ &minADE↓ & minFDE↓ & MR↓ \\
\hline
1  & 0.665 & 0.964 & 0.088 \\ 
3  & 0.663 & 0.952 & 0.085 \\
\rowcolor{gray!20}  
5  & 0.661 & 0.945 & 0.082 \\ 
7 & 0.664 & 0.950 & 0.083 \\
10 & 0.663 & 0.950 & 0.083 \\
\hline
\end{tabular}
\label{tab:sensitivity}
\end{table}
\subsection{Sensitivity Study}
The sensitivity study evaluates the impact of the hyperparameter $\lambda_1$ on model performance, which adjusts the relative weight of the second-stage loss. 
Through empirical studies, we investigate the effect of varying $\lambda_1$ around the experimental settings on the performance of the model. As shown in Table \ref{tab:sensitivity},  $\lambda_1$ is a sensitive parameter in our approach; setting $\lambda_1$ to 5 minimizes prediction errors in minADE, minFDE, and minMR, highlighting its significant impact on model performance. 

\subsection{Visualization}
The multi-modal prediction results of our method and the baseline HiVT-64 under different interaction intensity scenarios on the Argoverse 1 validation set are shown in Fig. \ref{fig:5}. In all three interaction scenarios—weak, moderate, and strong, our method consistently achieves more accurate final points of the best-predicted trajectory, indicated by the black rectangles, which are closer to ground-truth endpoints compared to HiVT-64. The visualization results demonstrate that our model exhibits clear advantages over the baselines in scenarios with moderate to strong interactions between agents.
In the weak interaction scenario, our method achieves higher accuracy.
In the moderate interaction scenario, LTMSformer achieves a reasonable turning radius, while the best-predicted trajectory of HiVT-64 deviates into the oncoming lane, resulting in an unreasonable turning radius.
In the strong interaction scenario, the predicted trajectories of HiVT-64 even exceed the lane boundaries. In contrast, LTMSformer stays within the lane boundaries, ensuring safer and more accurate predictions.
Overall, LTMSformer demonstrates higher prediction accuracy, achieves a reasonable turning radius, and maintains lane adherence.

\section{Conclusion}
\label{section5}
In this paper, we introduce a lightweight trajectory prediction framework, LTMSformer, which models temporal-spatial interaction features based on the proposed LTAA and MSE modules. LTAA captures the local temporal motion trends of agents, and MSE models spatial interactions by leveraging the high-order motion state attributes, ensuring that the predicted trajectories are both dynamically plausible and physically consistent. We also propose LPRM to refine initial trajectory predictions in a single pass, greatly improving prediction accuracy with fewer model parameters. Extensive experiments on the Argoverse 1 dataset demonstrate our superiority. In future work, we plan to extend our method to the Argoverse 2 dataset, which involves longer forecast horizons and more diverse dynamic objects.
\bibliographystyle{IEEEtran}
\bibliography{IEEE.bib}

\end{document}